\documentclass{article}

     \PassOptionsToPackage{numbers, compress}{natbib}

\usepackage[final]{neurips_2021}

\usepackage[utf8]{inputenc} 
\usepackage[T1]{fontenc}    
\usepackage[colorlinks=true, linkcolor=blue,citecolor=blue]{hyperref}       
\usepackage{url}            
\usepackage{booktabs}       
\usepackage{amsfonts}       
\usepackage{nicefrac}       
\usepackage{microtype}      
\usepackage{xcolor}         
\usepackage{amsmath}
\usepackage{amssymb}
\usepackage{graphicx}

\usepackage[pages=all,color=black]{background}

\newcommand\DupMark[1]{%
\textrm{Draft}\hspace{#1em}
     \textrm{Do not cite without permission of authors.}\hspace{#1em}
     \textrm{Draft}\hspace{#1em}
     \textrm{Do not cite without permission of authors.}\hspace{#1em}
     \textrm{Draft}\hspace{#1em}
}

\backgroundsetup{%
vshift=0pt,
hshift=150pt,
color=black,
  scale=2,       
  angle=45,       
  opacity=.02,    
  contents={%
\begin{minipage}{1\paperheight}
 \DupMark{3}\\[8ex]
  \DupMark{3}\\[8ex]
   \DupMark{3}\\[8ex]
    \DupMark{3}\\[8ex]
     \DupMark{3}\\[8ex]
      \DupMark{3}\\[8ex]
        \DupMark{3}\\[8ex]
     \DupMark{3}\\[8ex]
      \DupMark{3}\\[8ex]
      \DupMark{3}\\[8ex]
      \DupMark{3}\\[8ex]
 \end{minipage}
 }
 }

\title{Adversarial Attacks in Cooperative AI}

\author{%
  Ted Fujimoto \\
  Computing and Analytics Division\\
  Pacific Northwest National Laboratory\\
  Seattle, WA 98109 \\
  \texttt{ted.fujimoto@pnnl.gov} \\
  \And
  Arthur Paul Pedersen \\
  Department of Computer Science\\
  The City College of New York\\
  Grove School of Engineering\\
  City University of New York\\
  New York, NY 10010\\
  \texttt{apedersen@cs.ccny.cuny.edu} \\
}

\begin{document}

\maketitle

\begin{abstract} 
Single-agent reinforcement learning algorithms in a multi-agent environment are inadequate for fostering cooperation. If intelligent agents are to interact and work together to solve complex problems, methods that counter non-cooperative behavior are needed to facilitate the training of multiple agents. This is the goal of cooperative AI. Recent research in adversarial machine learning, however, shows that models (e.g., image classifiers) can be easily deceived into making inferior decisions.  Meanwhile, an important line of research in cooperative AI has focused on introducing algorithmic improvements that accelerate learning of optimally cooperative behavior. We argue that prominent methods of cooperative AI are exposed to weaknesses analogous to those studied in prior machine learning research. More specifically, we show that three algorithms inspired by human-like social intelligence are, in principle, vulnerable to attacks that exploit weaknesses introduced by cooperative AI's algorithmic improvements and report experimental findings that illustrate how these vulnerabilities can be exploited in practice.

\end{abstract}

\textcolor{black}{\textsc{Attention}: This is a preliminary draft of a working paper on ongoing research. Queries about the latest findings may directed to the authors. Do not cite this manuscript without permission of the authors.}
\section{Introduction}


All but assured in the era of ubiquitous AI is the concomitant ubiquity of interaction among intelligent systems and AI applications. This unshakable reality has led to a call for research in multi-agent reinforcement learning with a clear emphasis on cooperation \citep{dafoe2020open}. One motivation for this research is remedial: Single-agent reinforcement learning methods commonly promote the formation of selfish defectors within groups \citep{leibo2017multi}. No doubt would it be dangerous to deploy automated cars or planes which train on these methods. 
If intelligent agents are to interact with each other in the real world — the one we human beings inhabit, no less — cooperative AI is necessary to mitigate these potential dangers.

 

In machine learning, adversarial attacks are known to severely impair the performance of machine learning models by way of small manipulations to inputs at test time \citep{kurakin2017adversarial}. It has been shown, for example, that applying perturbations to an image can significantly decrease the accuracy of an image classifier, even when images and their perturbations are indistinguishable to the human eye. Aside from negatively affecting model performance, adversarial attacks can pose a hidden danger should they be utilized in a way that is not noticeable to a model's maintainer. Some researchers warn that there are currently no reliable defenses that cover a wide range of such attacks and accordingly argue against deploying machine learning solutions in contested, adversarial spaces \citep{short2019defending}. If cooperative AI is to improve human cooperation and well-being, a pressing question is whether, and to what extent, do analogous, new problems of adversarial attacks in machine learning subvert that benevolent goal. This paper endeavors to answer this pressing question.


An important line of research in cooperative AI has focused on introducing algorithmic improvements that accelerate learning of optimally cooperative behavior.  We show that three algorithms inspired by human-like social intelligence which have received much attention in multi-agent reinforcement learning are, in principle, vulnerable to attacks that exploit weaknesses introduced by cooperative AI's algorithmic improvements. We also report experimental findings that illustrate how these vulnerabilities can be exploited in practice.  Our findings show that cooperation could be subverted if, for example, an adversary makes a determination that cooperative AI techniques were used to train targets.


 Attacks on properties inspired by social intelligence have not been given their due in cooperative AI research. Carlini \citep{Carlini:website} reports over 6,000 research articles on adversarial supervised and single-agent reinforcement learning.  By contrast, we reckon there are fewer than one half dozen articles related to adversarial attacks in cooperative AI.



 



\subsection*{Outline}

This paper is organized as follows. Section \ref{Background:sec} argues that a dangerous form of social behavior, \emph{obfuscation}, demands urgent attention in cooperative AI.  Section \ref{RelatedWork:sec} places our line of research in the context of related research in cooperative AI, multi-agent reinforcement learning (\textsc{\textsc{marl}}), and adversarial machine learning. Section \ref{Methods:sec} introduces three new vulnerabilities corresponding to well-understood cooperative AI algorithms: public beliefs, mean-field representations, and sequential social dilemmas. Section \ref{ExperimentalResults:sec} thereupon presents experimental findings the show adversarial attacks negatively affect cooperative learning with public beliefs. The paper concludes with discussion of reported findings and ongoing work as well as proposals for future work in Section \ref{DiscussionAndFutureDirections:sec}.

The present manuscript reports the latest findings of ongoing research and is routinely updated at the conclusion of each experimental cycle.

\section{Background}\label{Background:sec}

The primary aim of cooperative AI, according to \citet{dafoe2020open}, is "to help individuals, humans and machines, to find ways to improve their joint welfare'' \citep[p. 4]{dafoe2020open}.  Nevertheless, its  methods, they caution, might fail to be socially beneficial, if left unchecked.   \citet{dafoe2020open} argue that if we do not take meaningful steps to ensure these methods maximize welfare, dangerous forms of behavior like exclusion, collusion, or coercion of humans or other agents might emerge. 

Omitted from among these potential risks, however, are forms of \emph{obfuscation} in cooperative behavior.  For example, a mediator facilitating negotiations over an issue of contention can, up to plausible deniability, amplify disagreement in order to sell a good whose value increases with longer runs of stalled negotiations.  Successful attacks could make well-trained cooperative AI systems perform much worse when deployed in the real world and cause catastrophic damage in critical applications, like self-driving vehicle coordination, AI-regulated marketplaces, or human-machine teaming in conflict resolution. Merely to suggest that cooperative AI's methods might not be socially beneficial would be to understate the risk of its menace to society. 





We investigate obfuscation in cooperative AI with a focus on scenarios featuring beliefs, mean-field approximations, and sequential social dilemmas, methods that have received a lot of attention from the cooperative AI community. These methods were chosen because they have been inspired by what people believe is necessary to intelligently navigate social interactions. For beliefs in \textsc{MARL}, our work investigates how adversarial attacks affect the public belief Markov decision process (\textsc{mdp}) \citep{foerster2019bayesian,sokota2021solving}. Reasoning about the beliefs of others helps us anticipate their actions. We also explore the adversarial vulnerabilities inherent in mean-field reinforcement learning \citep{carmona2019linear,yang2018mean}, which assumes that it is impractical to keep track of every single agent separately. In environments with a large number of agents, taking the average is more manageable. Sequential social dilemmas \citep{leibo2017multi} model how we choose to cooperate or defect over time. These scenarios are interesting because investigating obfuscations in these scenarios capture some of the strengths and weaknesses inherent in human cooperation.

 In Section \ref{ExperimentalResults:sec}, we apply adversarial attacks in machine learning as a way to model obfuscation in cooperative AI.

\section{Related Work}\label{RelatedWork:sec}

 Adversarial machine learning investigates the scenario where data is actively manipulated by an adversary seeking to make the model less accurate \citep{dalvi2004adversarial}. There has also been some more recent work specific to adversarial deep learning \citep{goodfellow2015,kurakin2017adversarial,szegedy2014}. Other work assumes an adversary in RL settings. \citet{huang2017adversarial} study the effects of white-box and black-box attack on neural network policies. \citet{rakhsha2020policy} provide a framework for training-time poisoning attacks on RL environments in offline and online settings. \citet{zhang2020adaptive} devise a framework and an efficient attack strategy for reward poisoning. While we build upon these previous works, we argue that cooperative AI introduces new vulnerabilities that have not been explored before in adversarial machine learning. Other research in cooperative AI draws attention to adversarial attacks on cooperative AI that particularly focus on multi-agent communication or consensus-based \textsc{MARL}. \citet{tu2021adversarial} explore the setting where RL agents communicate by sharing learned intermediate representations and how adversarial attacks diminish performance. By contrast, the methods we explore do not presume any explicit means of communication. \citet{figura2021adversarial} show that a single adversary can lead the other agents to optimize a utility function that it chooses. Our work makes the more general claim that, similar to computer vision \citep{short2019defending}, cooperative AI as a whole is vulnerable to adversarial attacks and research is needed to defend against such attacks. 

\section{Methods}\label{Methods:sec}

We now discuss some strategies an adversary might use to subvert cooperative AI. The following attacks are unique to cooperative AI and account for the interactions between agents. This non-exhaustive list of cooperative AI methods discusses possible attacks on beliefs, mean-field approximations, and inequalities in social dilemmas. Similar to how artificial neural networks were inspired by the brain, these cooperative AI methods were chosen because they were inspired by social concepts used to describe the facilitation of human and animal cooperation.

\subsection{Attacks on Beliefs}

In this paper, our primary focus will be on beliefs in multi-agent environments. In philosophy and game theory, the notion of belief has been explored in a way that coalesces with how we define knowledge and rationality (\citep{aumann1976agreeing}, \citep{pacuit2015epistemic}). It has also been investigated in the field of AI (\citep{gmytrasiewicz2005framework}, \citep{nayyar2013decentralized}). Beliefs are useful in \textsc{MARL} because it is a way for agents to infer the private features of the other agents. Here, we will focus on beliefs in terms of the public belief MDP (PuB-MDP) \citep{foerster2019bayesian}. In the PuB-MDP, beliefs are different from states or observations. Public beliefs are defined as the posterior over all private state features given the public state features: $\mathcal{B}_t = P(f^{pri} | f^{pub}_0 \dots f^{pub}_t)$, where $f^{pub}_t$ are the public features and $f^{pri}_t$ are the private features. Such beliefs are update with the following update rule:

\begin{equation}
    P(f^a_t | u^a_t, \mathcal{B}_t, f^{pub}_t, \hat{\pi}) = 
    \frac{P(u^a_t | f^a_t, \hat{\pi}) P(f^a_t | \mathcal{B}_t, f^{pub}_t)}{P(u^a_t | \mathcal{B}_t, f^{pub}_t, \hat{\pi})}
\end{equation}

where, at time $t$, $f^a_t$ are the private features of agent $a$, $u^a_t$ is the action of agent $a$, and $\hat{\pi}: \{f^a\} \rightarrow \mathcal{U}$ is a partial policy for the acting agent deterministically  mapping private observations to environment actions. 

The problem that arises is that even though this representation of beliefs can help groups of agents quickly learn to cooperative toward the optimal solution \citep{foerster2019bayesian,sokota2021solving}, it also opens up another vulnerability. Even if the designer of this multi-agent system made sure the states/observations could not be exploited, it is possible for an adversary to attack the belief representations to subvert the group of agents. Hence, the designer will need to account for another possible weakness in the multi-agent system. 

In the experiments section, we provide the results of some experiments inspired by past work on adversarial examples using FGSM \citep{goodfellow2015,huang2017adversarial} and poisoning rewards or certain features in RL during training as an adversarial attack to subvert the policy \citep{rakhsha2020policy, zhang2020adaptive}).

\subsection{Attacks on Mean-Field Approximations}

In cases where the number of agents is large, it is impractical for an RL algorithm to learn from all the observations and actions of each agent. In these cases, it is more practical to reduce all the observations or actions to a single value. Mean-field RL considers the scenario where the large number of agents prompts the need to reduce observations or actions to mean observation \citep{carmona2019linear} or mean actions \citep{yang2018mean} respectively. That is, for $N$ agents, each agent $i$ at time $t$ learns from its own observation $o^i_t$ and the mean observation of all agents $\frac{1}{N}\sum^N_{j=1} o^j_t$. For discrete actions, an agent $i$ can interpret the mean action as an empirical distribution of the actions the other agents take.

Although this method can make scaling a large number of agents more tractable, it can also introduce some vulnerabilities:

\begin{itemize}
    \item Negligibly small changes to each state of the neighboring agents can affect the mean observation. Typical attacks in single-agent RL make sure the attacks do not raise the attention of the designer so that the attacks do not stop. This is a trade-off since the attacker wants to inflict as much damage as possible. An attack's effect might be so small that it does not affect the victim at all. In large cooperative settings, however, the many small attacks can add up. If many small attacks were coordinated in a way that increases the bias of the sample mean state relative to the true mean state could harm the training of the mean-field RL agents. One possible strategy for the attacker is to apply Fast Gradient Sign Method (FGSM) \citep{goodfellow2015} to $n \le N$ agents. If the mean observation with $n$ many perturbations is not checked, then it might be that the effect becomes worse as $n$ gets closer to $N$.
    \item Keeping the mean action close to the uniform distribution. In most scenarios, having the policy be the uniform distribution at all states is not desirable. In particular, we assume that cooperation is reached when the group converges to predictable behavior for at least some of the states. An adversary could attack the mean action by modifying its empirical distribution to slightly put more weight on less optimal actions at each state.
\end{itemize}

Since adversarial attacks are effective when the attacks are not detected, precautions need to be taken so that attacks do not emerge from seemingly insignificant changes to individual observations and actions.  Currently, research on adversarial attacks in RL mostly describe the subversion of a single agent. There is a clear need to extend this to a larger, multi-agent setting.

\subsection{Attacks on Social Dilemma Inequalities}

Attacks on social dilemma inequalities are perhaps the most representative of what it means to subvert cooperation. In \citet{macy2002learning}, a set of inequalities were proposed to represent the social dilemmas of individual and collective interests. If we have variables $R$ (reward), $P$ (punishment), $S$ (sucker), and $T$ (temptation), then we have the following:

\begin{enumerate}
    \item $R > P$: Players prefer mutual cooperation over mutual defection.
    \item $R > S$: Players prefer mutual cooperation over unilateral cooperation.
    \item $2R > T + S$: Players prefer mutual cooperation over an equal probability of unilateral cooperation and defection.
    \item $T > R$: Players prefer unilateral defection to mutual cooperation (greed).\\
    or $P > S$: Players prefer mutual defection to unilateral cooperation (fear).
\end{enumerate}

\citet{leibo2017multi} coined the term \emph{sequential social dilemmas} (SSD) that extended these inequalities to a Markov game setting. To put it more clearly for a two player scenario, let $V^{\vec{\pi}}_i(s)$ be the state-value function for the joint policy of the two players: $\vec{\pi} = (\pi_1, \pi_2)$. Then the SSD can be modeled as a matrix game social dilemma (MGSD) for each state $s$ in the state space $\mathcal{S}$:

\begin{align}  
    R(s) &:= V^{\pi^C, \pi^C}_1(s) = V^{\pi^C, \pi^C}_2(s),\label{eq:2}\\ 
    P(s) &:= V^{\pi^D, \pi^D}_1(s) = V^{\pi^D, \pi^D}_2(s),\\
    S(s) &:= V^{\pi^C, \pi^D}_1(s) = V^{\pi^D, \pi^C}_2(s),\\
    T(s) &:= V^{\pi^D, \pi^C}_1(s) = V^{\pi^C, \pi^D}_2(s),\label{eq:5}
\end{align}

where $\pi^C$ and $\pi^D$ are cooperative and defecting policies. This set of value functions correspond to the variables in the social dilemma inequalities defined above. We also denote $\Pi^C$ as the set of all cooperative policies, and $\Pi^D$ as the set of all defecting policies.

Now that we have equations (\ref{eq:2}) - (\ref{eq:5}), we can better observe the weaknesses an adversary can exploit. If it is in our interest to cooperate, it is in our adversary's interest to subvert the chances of successful cooperation. One possible adversarial attack would be to affect the policies so that each agent is more likely to perceive unilateral defection from the other agents. In \citet{leibo2017multi}, choosing a cooperative or defecting policy depends on a \emph{social behavior metric} $\alpha: \Pi \rightarrow \mathbb{R}$. Let $\alpha_c$ and $\alpha_d$ be the threshold values such that $\alpha(\pi) < \alpha_c \Longleftrightarrow \pi \in \Pi^C$, and $\alpha(\pi) > \alpha_d \Longleftrightarrow \pi \in \Pi^D$. Hence, the social behavior metric $\alpha$ can be a vulnerability that allows an adversary to trick the agents into picking policies from $\Pi^D$ more often. One idealized, white-box attack an adversary could use is to train an RL agent with the reward function $r = 1$ if the victim agent fails to reach cooperative behavior and $r = 0$ otherwise at the end of an episode. The attack action space can be the interval $(-\delta, \delta)$ for some perturbation $\delta > 0$. The perturbation chosen by the attacker would then be applied to the victim's neural network policy. The attack's goal would then be to chose the sequence of perturbations that lead the victims to cross the social behavior metric and pick defecting policies as much as possible. This idealized attack draws inspiration from past adversarial RL work on attacks on neural network policies \citep{huang2017adversarial} and reward poisoning \citep{zhang2020adaptive}.

\section{Experimental Results}\label{ExperimentalResults:sec}

We have completed an experiment that shows how Fast Gradient Sign Method (FGSM) attacks \citep{goodfellow2015} affect the agents' beliefs. 

\subsection{Experiment Methodology}

We trained the agents using Cooperative Approximate Policy Iteration \citep{sokota2021solving} in the game of Trade Comm\footnote{Used the open-source code by \citet{sokota2021solving} here: \url{https://github.com/ssokota/capi} (MIT License)}. Trade Comm is a communication game with the following rules:

\begin{enumerate}
    \item Each player is independently dealt one of \texttt{num\_items} with uniform chance.
    \item Player 1 makes one of \texttt{num\_utterances} utterances, which is observed by player 2.
    \item Player 2 makes one of \texttt{num\_utterances} utterances, which is observed by player 1.
    \item Both players privately request one of the \texttt{num\_items} * \texttt{num\_items} possible trades.
\end{enumerate}

\begin{figure}[t]
    \centering
    \includegraphics[width=12cm, height=8cm]{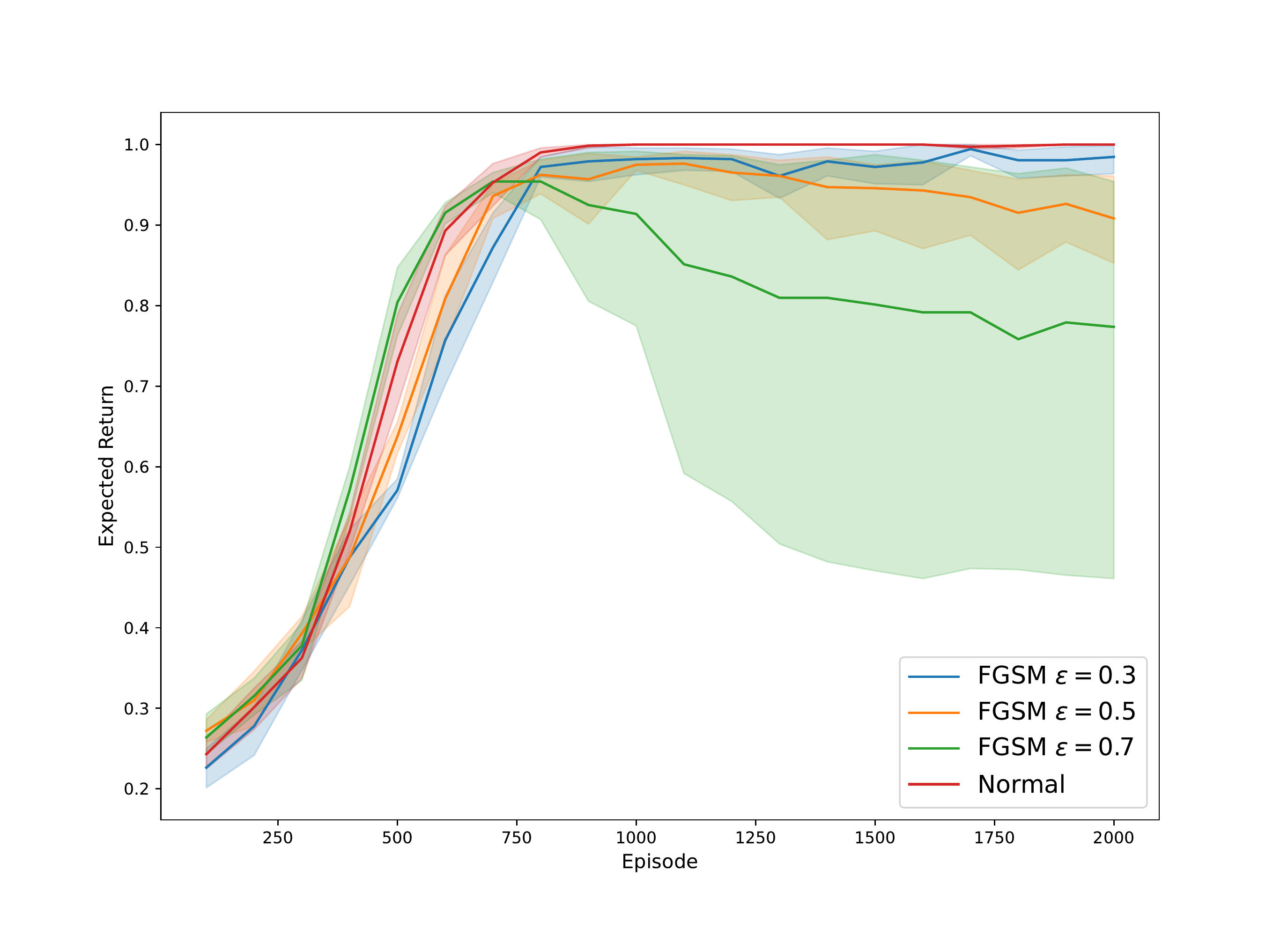}
    \caption{\footnotesize{The performance of four separate groups trained using Cooperative Approximate Policy Iteration. This plot shows the mean performance and 95\% confidence interval for each group's expected returns. As $\epsilon$ increases, the FGSM attack has more of an impact on the training.}}
    \label{fig:results}
\end{figure}

\begin{figure}[t]
    \centering
    \includegraphics[width=11cm, height=7cm]{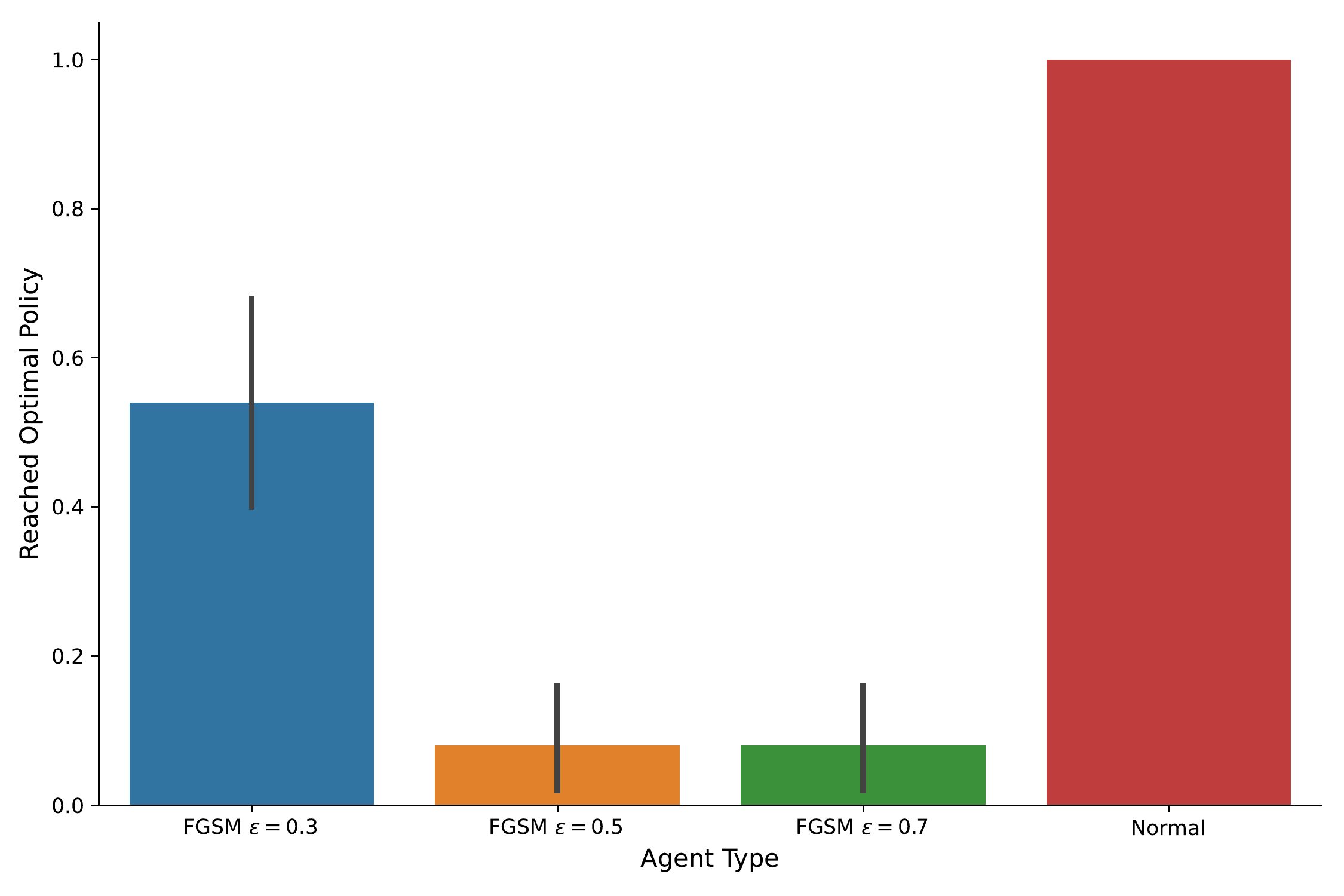}
    \caption{\footnotesize{This bar plot shows the percentage of times the Cooperative Approximate Policy Iteration groups achieved the optimal policy during training and 95\% confidence interval of the mean percentage. Here, we define ``reached optimal policy" as the expected returns being greater than 0.99.
    It shows how many times the groups reach the optimal policy after 1000 episodes, which is the halfway point.}}
    \label{fig:optimal_plot}
\end{figure}

The trade is successful if and only if both player 1 asks to trade its item for player 2's item and player 2 asks to trade its item for player 1's item. Both players receive a reward of one if the trade is successful and zero otherwise. We trained with 12 items and 12 utterances. The unperturbed belief representations are the updated belief distributions given the coordinator's prescription and the player's action. Hence, they are vectors with values in the interval $[0, 1]$. The public beliefs are represented as prescription vectors that are concatenations of the each player's posterior belief probabilities and the one-hot encoding of the players' utterances. During training, the CAPI algorithm generates a large number of prescription vectors using a policy, and then updates each vector's assessed according to their expected reward plus expected value of the next belief state. During evaluation, the policy selects the prescription with the highest assessed value at the current belief state as its action \citep{sokota2021solving}. One group was trained normally, and the other group was trained with adversarial perturbations applied to the agents' belief representations. This is a white-box attack since it has access to the parameters of the victim's policy. We trained each group over 5 random seeds. We used the same hyperparameters used in \citet{sokota2021solving}. The experiments were run on Amazon Web Services using NVIDIA Tesla V100 GPU or T4 Tensor Core GPU instances.

\subsection{Results}

From the plot in Figure \ref{fig:results}, it is clear that the perturbations hindered the training of the agents. The groups with adversarially perturbed beliefs do not converge to the optimally cooperative policy. On average, the normal group was able to converge after around 1000 episodes.  Our results show that FGSM can negatively affect the training of a cooperative group, and that increasing $\epsilon$ can proportionally impact training performance. The groups with perturbed by an FGSM attack with $\epsilon=0.5, 0.7$ have clearly worse performance than the normal group. Even the group with the smallest $\epsilon=0.3$ showed a statistically significant difference from the normal group. 

In the bar plots in Figure \ref{fig:optimal_plot}, we show the percentage of times the Cooperative Approximate Policy Iteration groups reach the optimal policy. Here, ``reaching the optimal policy" means the group achieves an expected return of at least 0.99 at the current time step during evaluation. After 1000 training episodes, the Normal group reaches the optimal policy close to 100\% of the time. The agents impacted by FGSM, however, show significant decreases in the frequency of reaching the optimal policy. Although the $\epsilon = 0.5$ and $\epsilon=0.7$ attacks have different mean expected returns in Figure \ref{fig:results}, they have similar frequencies in Figure \ref{fig:optimal_plot} because the bar plot does not account for the distance between the mean expected returns and the optimal score.

\section{Conclusion}\label{DiscussionAndFutureDirections:sec}

This paper has argued that prominent methods of cooperative AI are
exposed to systemic weaknesses analogous to those studied in prior machine learning
research on \emph{adversarial attacks}.   Forms of social behavior such as \emph{obfuscation} pose a danger to cooperative AI. Three well-understood algorithms --- public belief Markov decision processes, mean-field approximations, and multi-agent reinforcement learning in sequential social dilemmas --- are shown to be vulnerable to adversarial attacks adapted to their social environments.  Experimental findings are also offered to show that attacks on public beliefs can negatively impact convergence to optimally cooperative behavior that uses public beliefs.

\medskip

\small

\bibliographystyle{plainnat}
\bibliography{coop_bib}

\begin{thebibliography}{22}
\providecommand{\natexlab}[1]{#1}
\providecommand{\url}[1]{\texttt{#1}}
\expandafter\ifx\csname urlstyle\endcsname\relax
  \providecommand{\doi}[1]{doi: #1}\else
  \providecommand{\doi}{doi: \begingroup \urlstyle{rm}\Url}\fi

\bibitem[Aumann(1976)]{aumann1976agreeing}
Robert~J Aumann.
\newblock Agreeing to disagree.
\newblock \emph{The annals of statistics}, 4\penalty0 (6):\penalty0 1236--1239,
  1976.

\bibitem[Carlini(2022)]{Carlini:website}
Nicholas Carlini.
\newblock A complete list of all (arxiv) adversarial example papers, March
  2022.
\newblock URL
  \url{https://nicholas.carlini.com/writing/2019/all-adversarial-example-papers.html}.

\bibitem[Carmona et~al.(2019)Carmona, Lauri{\`e}re, and Tan]{carmona2019linear}
Ren{\'e} Carmona, Mathieu Lauri{\`e}re, and Zongjun Tan.
\newblock Linear-quadratic mean-field reinforcement learning: convergence of
  policy gradient methods.
\newblock \emph{arXiv preprint arXiv:1910.04295}, 2019.

\bibitem[Dafoe et~al.(2020)Dafoe, Hughes, Bachrach, Collins, McKee, Leibo,
  Larson, and Graepel]{dafoe2020open}
Allan Dafoe, Edward Hughes, Yoram Bachrach, Tantum Collins, Kevin~R McKee,
  Joel~Z Leibo, Kate Larson, and Thore Graepel.
\newblock Open problems in cooperative ai.
\newblock \emph{arXiv preprint arXiv:2012.08630}, 2020.

\bibitem[Dalvi et~al.(2004)Dalvi, Domingos, Sanghai, and
  Verma]{dalvi2004adversarial}
Nilesh Dalvi, Pedro Domingos, Sumit Sanghai, and Deepak Verma.
\newblock Adversarial classification.
\newblock In \emph{Proceedings of the tenth ACM SIGKDD international conference
  on Knowledge discovery and data mining}, pages 99--108, 2004.

\bibitem[Figura et~al.(2021)Figura, Kosaraju, and Gupta]{figura2021adversarial}
Martin Figura, Krishna~Chaitanya Kosaraju, and Vijay Gupta.
\newblock Adversarial attacks in consensus-based multi-agent reinforcement
  learning, 2021.

\bibitem[Foerster et~al.(2019)Foerster, Song, Hughes, Burch, Dunning, Whiteson,
  Botvinick, and Bowling]{foerster2019bayesian}
Jakob Foerster, Francis Song, Edward Hughes, Neil Burch, Iain Dunning, Shimon
  Whiteson, Matthew Botvinick, and Michael Bowling.
\newblock Bayesian action decoder for deep multi-agent reinforcement learning.
\newblock In \emph{International Conference on Machine Learning}, pages
  1942--1951. PMLR, 2019.

\bibitem[Gmytrasiewicz and Doshi(2005)]{gmytrasiewicz2005framework}
Piotr~J Gmytrasiewicz and Prashant Doshi.
\newblock A framework for sequential planning in multi-agent settings.
\newblock \emph{Journal of Artificial Intelligence Research}, 24:\penalty0
  49--79, 2005.

\bibitem[Goodfellow et~al.(2015)Goodfellow, Shlens, and
  Szegedy]{goodfellow2015}
Ian Goodfellow, Jonathon Shlens, and Christian Szegedy.
\newblock Explaining and harnessing adversarial examples.
\newblock In \emph{International Conference on Learning Representations}, 2015.
\newblock URL \url{http://arxiv.org/abs/1412.6572}.

\bibitem[Huang et~al.(2017)Huang, Papernot, Goodfellow, Duan, and
  Abbeel]{huang2017adversarial}
Sandy Huang, Nicolas Papernot, Ian Goodfellow, Yan Duan, and Pieter Abbeel.
\newblock Adversarial attacks on neural network policies.
\newblock \emph{arXiv preprint arXiv:1702.02284}, 2017.

\bibitem[Kurakin et~al.(2017)Kurakin, Goodfellow, and
  Bengio]{kurakin2017adversarial}
Alexey Kurakin, Ian Goodfellow, and Samy Bengio.
\newblock Adversarial machine learning at scale.
\newblock In \emph{International Conference on Learning Representations}, 2017.

\bibitem[Leibo et~al.(2017)Leibo, Zambaldi, Lanctot, Marecki, and
  Graepel]{leibo2017multi}
Joel~Z Leibo, Vinicius Zambaldi, Marc Lanctot, Janusz Marecki, and Thore
  Graepel.
\newblock Multi-agent reinforcement learning in sequential social dilemmas.
\newblock In \emph{Proceedings of the 16th Conference on Autonomous Agents and
  MultiAgent Systems}, pages 464--473, 2017.

\bibitem[Macy and Flache(2002)]{macy2002learning}
Michael~W Macy and Andreas Flache.
\newblock Learning dynamics in social dilemmas.
\newblock \emph{Proceedings of the National Academy of Sciences}, 99\penalty0
  (suppl 3):\penalty0 7229--7236, 2002.

\bibitem[Nayyar et~al.(2013)Nayyar, Mahajan, and
  Teneketzis]{nayyar2013decentralized}
Ashutosh Nayyar, Aditya Mahajan, and Demosthenis Teneketzis.
\newblock Decentralized stochastic control with partial history sharing: A
  common information approach.
\newblock \emph{IEEE Transactions on Automatic Control}, 58\penalty0
  (7):\penalty0 1644--1658, 2013.

\bibitem[Pacuit and Roy(2015)]{pacuit2015epistemic}
Eric Pacuit and Olivier Roy.
\newblock Epistemic foundations of game theory.
\newblock \emph{Stanford Encyclopedia of Philosophy}, 2015.

\bibitem[Rakhsha et~al.(2020)Rakhsha, Radanovic, Devidze, Zhu, and
  Singla]{rakhsha2020policy}
Amin Rakhsha, Goran Radanovic, Rati Devidze, Xiaojin Zhu, and Adish Singla.
\newblock Policy teaching via environment poisoning: Training-time adversarial
  attacks against reinforcement learning.
\newblock In \emph{International Conference on Machine Learning}, pages
  7974--7984. PMLR, 2020.

\bibitem[Short et~al.(2019)Short, La~Pay, and Gandhi]{short2019defending}
Austin Short, Trevor La~Pay, and Apurva Gandhi.
\newblock Defending against adversarial examples.
\newblock Technical report, Sandia National Lab.(SNL-NM), Albuquerque, NM
  (United States), 2019.

\bibitem[Sokota et~al.(2021)Sokota, Lockhart, Timbers, Davoodi, D'Orazio,
  Burch, Schmid, Bowling, and Lanctot]{sokota2021solving}
Samuel Sokota, Edward Lockhart, Finbarr Timbers, Elnaz Davoodi, Ryan D'Orazio,
  Neil Burch, Martin Schmid, Michael Bowling, and Marc Lanctot.
\newblock Solving common-payoff games with approximate policy iteration.
\newblock In \emph{Proceedings of the AAAI Conference on Artificial
  Intelligence}, volume~35, pages 9695--9703, 2021.

\bibitem[Szegedy et~al.(2014)Szegedy, Zaremba, Sutskever, Bruna, Erhan,
  Goodfellow, and Fergus]{szegedy2014}
Christian Szegedy, Wojciech Zaremba, Ilya Sutskever, Joan Bruna, Dumitru Erhan,
  Ian Goodfellow, and Rob Fergus.
\newblock Intriguing properties of neural networks.
\newblock In \emph{International Conference on Learning Representations}, 2014.
\newblock URL \url{http://arxiv.org/abs/1312.6199}.

\bibitem[Tu et~al.(2021)Tu, Wang, Wang, Manivasagam, Ren, and
  Urtasun]{tu2021adversarial}
James Tu, Tsunhsuan Wang, Jingkang Wang, Sivabalan Manivasagam, Mengye Ren, and
  Raquel Urtasun.
\newblock Adversarial attacks on multi-agent communication.
\newblock \emph{arXiv preprint arXiv:2101.06560}, 2021.

\bibitem[Yang et~al.(2018)Yang, Luo, Li, Zhou, Zhang, and Wang]{yang2018mean}
Yaodong Yang, Rui Luo, Minne Li, Ming Zhou, Weinan Zhang, and Jun Wang.
\newblock Mean field multi-agent reinforcement learning.
\newblock In \emph{International Conference on Machine Learning}, pages
  5571--5580. PMLR, 2018.

\bibitem[Zhang et~al.(2020)Zhang, Ma, Singla, and Zhu]{zhang2020adaptive}
Xuezhou Zhang, Yuzhe Ma, Adish Singla, and Xiaojin Zhu.
\newblock Adaptive reward-poisoning attacks against reinforcement learning.
\newblock In \emph{International Conference on Machine Learning}, pages
  11225--11234. PMLR, 2020.

\end{thebibliography}

\end{document}